\documentclass[a4paper,conference]{IEEEtran}

\usepackage{enumerate}
\usepackage{amsmath}
\usepackage[printwatermark]{xwatermark}
\usepackage{xcolor}

\begin{document}

\newwatermark[pagex={1},fontfamily=bch,fontsize=11pt,color=gray!60,angle=0,scale=1,xpos=0,ypos=14cm]{This paper has been accepted for publication\\ at the IAPR IEEE/Computer Society International Conference on Pattern Recognition (ICPR), Milan, 2021.}

%
\title{Attention-Driven Body Pose Encoding for Human Activity Recognition}

\author{\IEEEauthorblockN{Bappaditya Debnath\IEEEauthorrefmark{1},
Swagat Kumar\IEEEauthorrefmark{1},
Mary O'Brien\IEEEauthorrefmark{2} and
Ardhendu Behera\IEEEauthorrefmark{1}}
\IEEEauthorblockA{\IEEEauthorrefmark{1}Department of Computer Science, \IEEEauthorrefmark{2} Faculty of Health, Social Care \& Medicine\\
Edge Hill University, Ormskirk, UK, L394QP\\ 
Email: \{debnathb, obrienm, kumars, beheraa\}@edgehill.ac.uk}}


%

\maketitle

\begin{abstract}
This article proposes a novel attention-based body pose encoding for human activity recognition that presents a enriched representation of body-pose that is learned.
The enriched data complements the 3D body joint position data and improves model performance. In this paper, we propose a novel approach that learns enhanced feature representations from a given sequence of 3D body joints. To achieve this encoding, the approach exploits 1) a spatial stream which encodes the spatial relationship between various body joints at each time point to learn spatial structure involving the spatial distribution of different body joints 2) a temporal stream that learns the temporal variation of individual body joints over the entire sequence duration to present a temporally enhanced representation. Afterwards, these two pose streams are fused with a multi-head attention mechanism. 
We also capture the contextual information from the RGB video stream using a Inception-ResNet-V2 model combined with a multi-head attention and a bidirectional Long Short-Term Memory (LSTM) network. 
Finally, the RGB video stream is combined with the fused body pose stream to give a novel end-to-end deep model for effective human activity recognition. 
\end{abstract}

%
\IEEEpeerreviewmaketitle

\section{Introduction}
Human activity recognition from videos has several potential applications such as home-based rehabilitation, elderly monitoring, human-human interaction, and so on. Therefore, it has received considerable attention from the computer vision community. Researchers have used many techniques ranging from simple tracked key-points \cite{messing2009activity, wang2012mining, eweiwi2014efficient} to state-of-the-art deep CNN networks \cite{Baradel_2018_CVPR, shahroudy2017deep, zhang2017view, song2017end}. However, 
it is still an unsolved problem and is mainly due to the difficult nature of human movements. The difficulty is often influenced by many factors such as wide variations in executing a given activity, different environment conditions (e.g. background scene, lighting, etc.), unavoidable occlusions, intra-class variations, and similarity between various activities. Research in this area has significantly benefited from the recent advances in deep learning models such as Temporal Convolutional Networks (TCN) \cite{kim2017interpretable}, attention mechanisms \cite{song2017end}, and so on. With the availability of cheap commercial devices such as Kinect, both RGB videos and human body skeleton represented by 3D body joints are readily available. Moreover, the availability of large scale datasets \cite{shahroudy2016ntu} and having both RGB-D and skeleton information have significantly contributed in advancing the field. However, owing to the challenges mentioned above, activity recognition remains an active area of research. 

For human activity recognition authors have developed models, which explore RGB video data, body pose information, depth data and/or various combination of these data types \cite{shahroudy2016ntu, shahroudy2017deep,Baradel_2018_CVPR,shahroudy2015multimodal}. The RGB video data is often combined with pose information to take advantage of information contained in both types of data \cite{baradel2018human, shahroudy2017deep, Baradel_2018_CVPR}. In this article, we propose a novel attention-based multi-stream deep architecture that combines video frame data and human body pose information. The proposed model outperforms the state-of-the-art approaches on three challenging datasets. Typically, pose information consists of human joint positions in 2D/3D and is provided for each frame. Researchers have tried to improve the activity recognition performance by encoding additional information such as velocity, acceleration and pairwise relationships involving various body joints. Instead of encoding additional handcrafted pose-related features, the proposed attention-driven body pose network learns such encodings. Our pose network consists of a Spatial Encoding Unit (SEU) and a Temporal Encoding Unit (TEU). The SEU provides an enriched representation that \textit{learns to capture} the structural relationship between various body joints at each frame in a given sequence. This presents a spatially enhanced representation of the skeleton sequence to the network and is learned not hand-crafted. On the other hand, the TEU encodes the temporal relationship of each body joint over the duration of a given sequence to present a temporally enhanced representation of the pose sequence.

\begin{figure*}[h]
\centering
\includegraphics[width=0.7\textwidth]{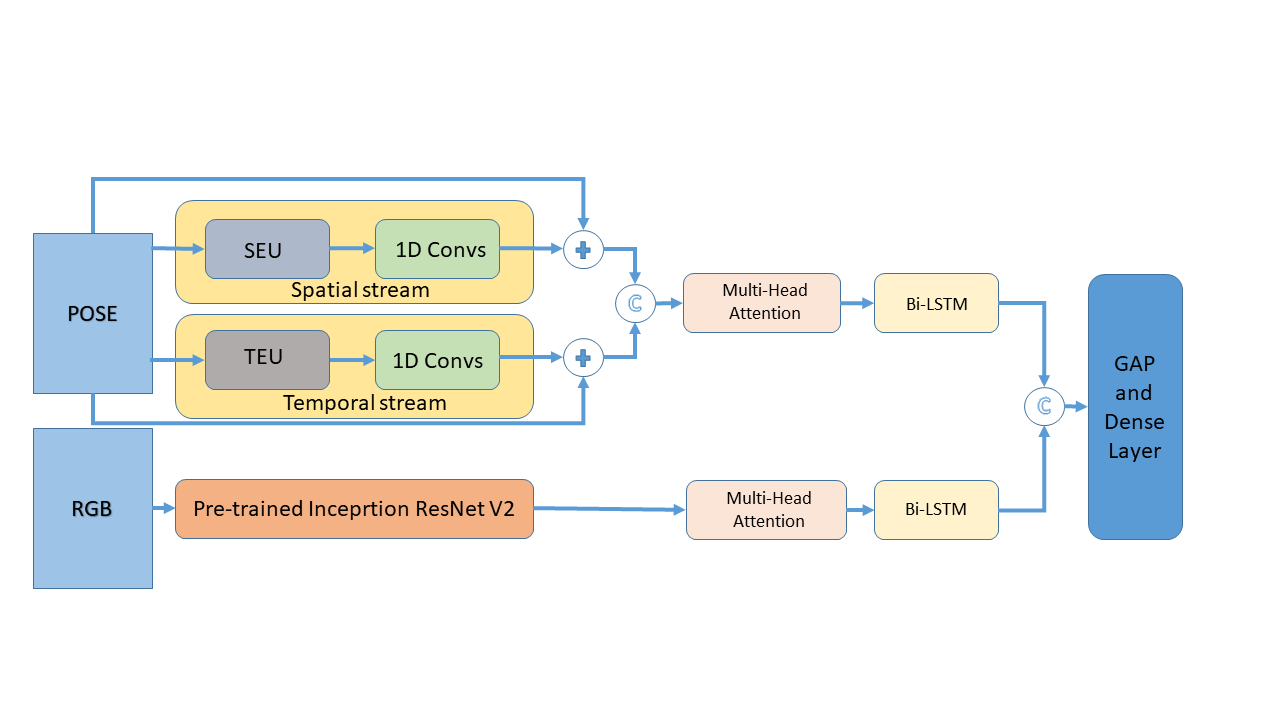}
\caption{We propose a novel skeleton sequence encoding approach. The Spatial Encoding Unit (SEU) learns the structural dependencies and relationships between various body joints and presents a spatially enhanced sequence to the network. The Temporal Encoding Unit (TEU) learns the frame-wise position of each joint to learn a temporally augmented meaningful representation. The `+' symbol stands for addition while `C' indicates concatenation}
\label{fig:model}
\vspace{-2mm}
\end{figure*}

Many existing researches suggest that combining multi-modal information (e.g. RGB, skeletal, depth) has outperformed the models that rely on single modality \cite{baradel2018human, shahroudy2017deep, Baradel_2018_CVPR}. For example, RGB data can provide contextual scene cues such as background, motion, texture etc. On the other hand, pose information provides a sequence of 2D/3D coordinates that the network needs to focus on. The following key points should be considered while designing and developing an efficient model that combines RGB and skeletal information:
\begin{enumerate}[i]
    \item Video sequences contain a high amount of visual, as well as temporal information. Deep CNN models are very good at capturing visual information, but they are unable to encode the temporal information contained in video sequences. 
    \item Pose sequence-based models should be able to capture the long-term temporal dependencies.    
    \item Pose-based models should also be able to learn the spatial relationships between various body-joints in order to semantically encode the structural relationships and various inter-dependencies among different body parts. 
\end{enumerate}
Modelling the above-mentioned main points and combining multi-modal information in a meaningful way is the key to overcome some of the key challenges in human activity recognition problem. The proposed model aims to address this by developing an end-to-end deep architecture consisting of two skeleton and one RGB stream as shown in Fig. ~\ref{fig:model}. For the RGB stream, a pre-trained Inception-ResNet-V2 \cite{szegedy2017inception} is used to process every frame in a RGB video in a time distributed manner. This is followed by a Self-Multi-Head Attention mechanism and a bidirectional LSTM (bi-LSTM). While the pre-trained network effectively captures the spatial information, the bi-LSTM learns to capture the temporal information in videos. The Multi-Head Attention mechanism further enhances the network performance by focusing on the visual features that are important for discrimination. The skeleton information is processed through a spatial stream and a temporal stream in parallel. The streams are then concatenated and passed through a Self-Multi-Head Attention mechanism followed by a bi-LSTM. Generally, in pose-base models, each joint is represented by a vector of length three representing its 3D positions (x, y and z).  Sometimes, this is enriched with other information such as distance between two body joints, pairwise relations which presents an augmented and enriched vector for each joint. Instead, the SEU learns such representations through 1D convolutions. The SEU processes joint information in each frame separately and thus captures the spatial structural relationship between various body joints. On the other hand, the TEU learns the dynamics of these body joints instead of handcrafted augmented information such as velocity and acceleration and captures the long-term temporal dependencies in a given pose sequence. 
Our contributions are:
\begin{enumerate}
	\item We present a novel architecture that encodes the structural relationships and dependencies between various body joints, as well as captures long-term temporal dependencies of each body joint.
	\item We present a novel attention-based approach that \textit{learns to attend} different parts of a feature vector corresponding to a given stream, and combines them in an efficient way. The deep CNN and 1D convolution sub-network benefits from both RGB and pose information to give us state-of-the-art results across three datasets including the challenging NTU-RGBD dataset.  
\end{enumerate}
\section{Related Work}
Our work focuses on three main components: 1) human activity recognition, 2) human body pose representation for activity modeling and recognition, and 3) attention mechanisms in improving activity recognition performance. Therefore, in this section, we revisit the existing researches covering these three main areas:\\
\textbf{Activity Recognition: } Traditional approaches involving human activity recognition have mainly focused on monocular RGB video data \cite{herath2017going}. In order to process RGB+Time data, stacked sequences of frames are encoded through 2D CNN in a time distributed manner \cite{ma2016going,deng2016structure} or processed through 3D CNN \cite{wu2016deep,molchanov2016online,ji20123d}. The 3D convolution for activity recognition was first introduced by Tran et al. \cite{tran2015learning}. Often, researchers have combined multiple streams to boost activity recognition performance. Ma et al. \cite{ma2016going} proposed a three-stream network, where two of the streams focus on regions of interest while the third stream concentrates on the optical flow. Similarly, Deng et al. \cite{deng2016structure} combine an activity CNN with a scene CNN to recognize group activities. Baradel et al. \cite{Baradel_2018_CVPR} use an attention-based interest point called glimpse clouds involving ResNet-50. Molchanov et al. advocate recurrent 3D CNN \cite{molchanov2016online} for online detection of hand gestures. Sharma et al.\cite{sharma2015action} propose a model that integrates features from different parts of a spatiotemporal LSTM network and makes soft attention-based decision to recognize activities. 

Images or videos from monocular cameras do not contain depth information. With the availability of cheap depth sensors such as Microsoft Kinect \cite{han2013enhanced}, depth information is now readily available. As a result, 3D pose information extracted from depth-enabled devices like Kinect have added as another modality for activity recognition. Thus, many recent human activity recognition methods often combine RGB+Time with skeleton sequence data. Body pose data are normally available as 3D joint positions and are often processed using recurrent networks. 
Such networks are in the form of Recurrent Neural Network (RNN) or its extensions such as LSTM or GRU. Recently, Temporal Convolutional Network (TCN) has been explored for pose sequence processing and analysis \cite{kim2017interpretable}. TCNs can be seen as a 1D fully connected network combined with causal convolution \cite{lea2017temporal}. The size of the data involving 3D body poses is significantly less than the image data and therefore, to increase the recognition accuracy, different handcrafted data augmentation techniques are explored by various researchers. Some of these are: augmenting coordinates with velocities and acceleration \cite{demisse2018pose,zanfir2013moving}, various normalization techniques for the body joints \cite{zanfir2013moving}, and relative positions \cite{ke2017new}. Instead of handcrafting features for enhancing representation, the purpose of the proposed spatial and temporal encoding is to automatically learn representations that can contribute towards performance.\\ 
\textbf{Human Body Pose Models: }
Pose-base models attempt to learn the structural information by capturing the various inter-joint relationships and dependencies and learn to recognize how these representations vary over time for various activity classes. Wang et al. \cite{wang2012mining} advocated a method that splits body joints into five groups and then use spatial and temporal dictionaries to encode the spatial structure of human bodies. Vemulapalli et al. \cite{vemulapalli2014human} consider the affine transformations to represent geometric relationships of body parts through Lie groups. The authors have extensively used RNNs for representing the skeleton sequences. Similarly, Du et al. \cite{du2015hierarchical} have used RNNs in a hierarchical manner to represent groups of body joints. Each joint is represented by a sub-network at the initial layer, then the joint representations are fused hierarchically to form groups of joints. Similarly, Shahroudy et al. \cite{shahroudy2016ntu} have used body part-aware LSTM networks for encoding skeleton sequences. In \cite{kim2017interpretable}, the authors have considered TCN with residual connections for pose-based interpret-able activity recognition. In our proposed model, 1d convolution is used to process the pose sequence. The proposed approach is inspired by the works of Kim et al. \cite{kim2017interpretable} and Xu et al. \cite{xu2018ensemble} who have used TCN with residual connections. The authors have considered an ensemble of spatial-temporal, hierarchical and attention-based networks to boost the performance of their network. Similarly, Song et al. \cite{song2017end} have introduced a separate spatial and temporal attention-based networks for skeleton-based activity recognition. For each frame, the spatial network attaches more weight to joints that are important to the current activity. Whereas, the temporal network selects the more important frames.\\
\textbf{Attention mechanism: }
Attention mechanisms are used to selectively focus on more relevant and discriminatory features \cite{bahdanau2014neural} and is inspired by the selective search, which is commonly appeared in human visual system. The mechanism calculates similarity between input vectors ``queries'' and ``keys'', and then maps input ``values'' to output vectors based on this calculated similarity. In this case, queries, keys and values are all vectors. 
Zhang et al. \cite{zhang2018self} proposed  a self-attention mechanism that relates various temporal position of the same sequence to calculate a weighted representation of itself. Therefore, in self-attention queries, keys and values are the same vector. Recently, Multi-Head attention mechanism \cite{vaswani2017attention} has been successfully used in sequence modeling. As the name suggests, Multi-Head Attention models linearly juxtapose output of scaled dot product attention into number of groups (heads). This allows the model to represent different learned sub-spaces at different positions. Attention mechanisms have been widely adopted for image and video understanding tasks 
\cite{cho2015describing,xu2015show,sharma2015action,song2017end,jaderberg2015spatial}. 
In our proposed method, Multi-Head attention \cite{vaswani2017attention} is adapted to improve the discriminatory capabilities involving visual and pose feature maps extracted using the respective deep CNN focusing on video frames and our proposed SEU and TEU modules (see Fig. \ref{fig:model}) for encoding skeletal data. 
\section{Proposed Approach}
The architecture of the proposed network is shown in Fig. \ref{fig:model}. The model takes input as a video sequence and body pose sequence, and provides out as an activity class label to the input sequence. The model introduces a novel two-stream attention-based joint position encoding framework that temporally and spatially learn the structural relationships between various body joints. The feature maps describing the spatial and temporal structures are concatenated in the final representation. Afterwards, the concatenated skeleton stream is combined with an attention-based time distributed CNN network in a late fusion mode (Fig. \ref{fig:model}). From the literature review, we observe that instead of presenting sequences of joints directly to the network, the authors have tried to learn more enriched representations. For example, grouping of joints through a hierarchical network \cite{wang2012mining}. Other methods include enriching the representations by presentation of hand-crafted features \cite{vemulapalli2014human}, whereas the proposed network automatically learn these representations. Generally, a sequential network such as TCN or RNN only learns the temporal relationship between frames. On the other hand, our proposed network i) learns the structural relationships between various body joints, and ii) learns the frame-wise relationship of each joint, in addition to learning the temporal relationships between frames. In the following subsections, we elaborate the spatial and temporal encoding units, the RGB model and the attention mechanism used in our model.
\subsection{Human Body Pose: Spatial Stream}
\begin{figure}[ht]
\centering
\includegraphics[width=0.45\textwidth]{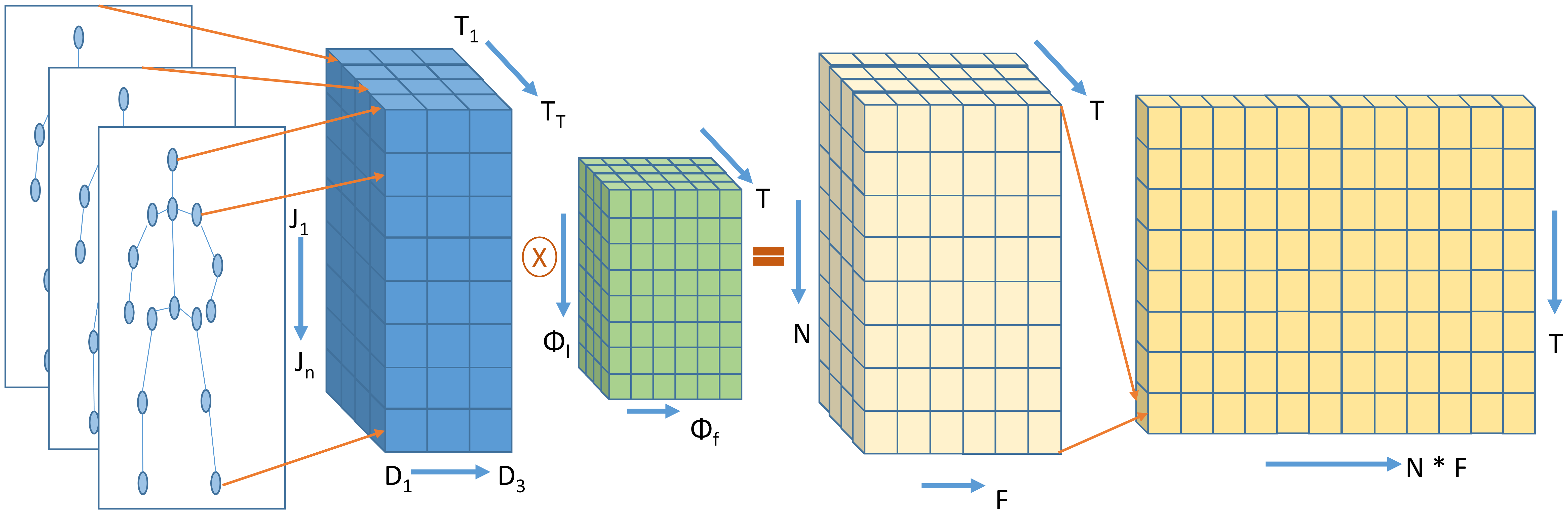}
\caption{The Spatial Encoding Unit (SEU) augments the spatial data with learned representations. Typically, a matrix of size $T,N*D$ is presented for sequential processing, instead, we present a learned representation of size $T,N*F$. $T$ is time or number of frames, $F$ is the number of filters and $N$ is the number of joints. $D$ is normally 3 (x, y, z) representing 3D positions, and is often enhanced with additional handcrafted features such as velocity, acceleration etc. Instead, the SEU learns $F$ representations per $N$ joints per $T$ time-steps. The `X' symbol indicates convolution}
\label{fig:SEU}
\end{figure}

The spatial stream consists of a Spatial Encoding Unit (SEU) followed by three layers of 1D convolutions. The goal of the SEU is to present enriched pose information to the network for better performance. Normally, for sequential processing, the input consists of 2D or 3D joint coordinates for each frame. To present a richer representation, we use a three layer 1D convolutional network which learns the structural information between various body joints. Let's consider there are $T$ number of frames in a sequence, $J$ is the number of body joints and $F$ is the total number of filters in a layer. An 1D convolution operation performs the following mapping with input vector V $\in \mathbf{R}^{F \times {JD}}$ :
\begin{equation}
    M_{T,F}  = U(\Theta_{F}, V_{T, J*D})
    \label{eq:eq1}
\end{equation}
where $U$ is the convolution operation parametrized by filters $\Theta_{F}$ and $D$ indicates the number of dimensions of each body joint and in our case, it is 3D positional information (x, y, z) or more if we have additional handcrafted representations. For each frame $t \in \{1\dots T\}$ in the sequence, we encode pose vector $V \in \mathbf{R}^{J\times D}$ through 1D convolution operations. Formally, we perform the following operation:
\begin{equation}
  M^t_{J,F}  = U_{t}(\Theta_{F}, V_{J,D})\\
  \label{eq:eq2}
\end{equation}
\begin{equation}
    M^t_{J, F} \rightarrow M_{T, J*F}
    \label{eq:eq3}
\end{equation}
As shown in Eq. \ref{eq:eq2}, for each time step or frame, we perform a convolution operation where each joint is represented individually. The learned map is then spatially squeezed and aggregated temporally as shown in Fig. \ref{fig:SEU} (Eq. \ref{eq:eq3}). Normally, while encoding skeleton sequence, a 2D vector of dimensions ($T,N*3$) is presented to the network. Instead, we present a learned representation of size ($T, J*F$). For each joint $j$ in every frame at position $t\in \{1\dots T\}$, we have $F$ filters representing the learned encoding. Whereas, in normal practice, it is often enriched the skeleton sequence with handcrafted mechanisms such as groups of joints \cite{wang2012mining}, velocity and acceleration \cite{zanfir2013moving} and so on. However, in our case, we learn such representations. This enriched representation is presented to the spatial stream which consists of 3 layers of 1D convolutions.  
\subsection{Human Body Pose: Temporal Stream}
\begin{figure}[ht]
\centering
\includegraphics[width=0.45\textwidth]{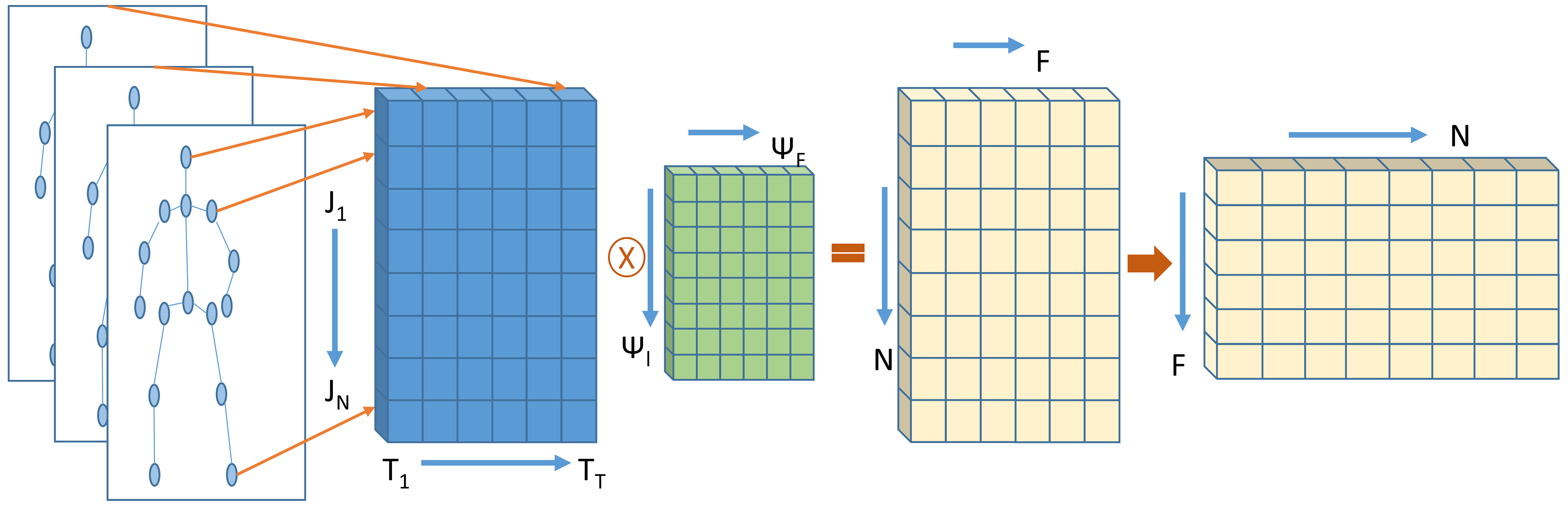}
\caption{The Temporal Encoding Unit (TEU) encodes frame-wise positions of each individual body joint to learn temporally augmented representations. Instead of temporal length $T$ we present an learned temporal sequence of length $F$ determined by the number of filters. The `X' symbol indicates convolution}
\label{fig:TEU}
\end{figure}
Similar to the spatial stream, the temporal stream consists of the TEU followed by three layers of 1D convolutions. The goal of the TEU is to encode the frame-wise positions of body joints and present a temporally enhanced representation for each joint, individually. Similar to the SEU, we use three layers of 1D convolutions. For each joint $j \in \{1\dots J\}$, we encode a vector $\bar{V} \in \mathbf{R}^{J\times{T}}$, through 1D convolutions parametrized by filters $\Psi_{F}$. Formally, we perform the following transformation:
\begin{equation}
  \bar{M}_{J, F}  = \bar{U}(\Psi_{F}, \bar{V}_{J,T})
  \label{eq:eq4}
\end{equation}
\begin{equation}
  \bar{M}_{J, F} \rightarrow \bar{M}_{F, J}
  \label{eq:eq6}
\end{equation}
If compared to the Eq. \ref{eq:eq1}, we get a feature map of dimensions ($F, J$) instead of ($T, J*F$). Thus, as shown in Fig. \ref{fig:TEU}, our TEU represents a map with temporal size $F$ instead of  $T$. This is equivalent to augmenting the temporal dimension of the input vector from ${T}$ to ${F}$ based on the number of filters. Enhancing the temporal dimension in such a manner benefits the network. The enhanced temporal representation is fed into the temporal stream, which consists of three layers of 1D convolutions.  
\subsection{Human Body Pose: Fusion of Spatial and Temporal Stream}
The impact of residual connection is well-studied \cite{he2016deep} and the proposed network also benefits from residual connections. The resulting maps from both the temporal and spatial streams are added to residual connections as shown in Fig. \ref{fig:model}. Ba et al. \cite{ba2016layer} argued that for sequential networks, layer normalization is beneficial when compared to batch normalization. We observe that our model also benefits from the layer normalization, which is added to the residual connections. The streams are then concatenated along the time axis. As a result, the fused pose stream has a compact yet richer representation of both the spatial and the temporal characteristics of the given sequence. In order to exploit this representation, we further used Multi-Head Attention with LSTM as shown in Fig \ref{fig:model}. The attention mechanism learns weighted representations of different temporal sub-spaces to focus the network on important temporal zones for discrimination. The representations of the attention-based fused pose stream is very rich and is different from the individual stream. Therefore, it is beneficial to exploit the same with further processing based on bidirectional LSTM to capture the long-term sequential information.    
\subsection{Context/scene Descriptor: Processing of RGB Stream}
Unlike in body pose representation using skeletal information consisting of 3D positions of body joints, video frames provide richer information. These information are explored to describe the contextual/scene descriptor. Our proposed approach is no different to it. As shown in Fig. \ref{fig:model}, we use a pre-trained Inception-Resnet-V2 \cite{szegedy2017inception} model to extract this contextual scene descriptor. The model is used in a time distributed manner i.e. the same model is used to extract the contextual descriptor for each frame in a video sequence. 
The Inception-Resnet-V2 is a well-known network which combines the advantages of inception modules \cite{szegedy2015going} with residual connections \cite{he2016deep}. While CNN models are very good at visual recognition tasks (e.g. image recognition), it is often required to have additional temporal processing \cite{okita2017recognition} for activity recognition tasks involving video sequences. Recurrent mechanisms like Recurrent Neural Networks (RNNs) and LSTMs are suitable for such tasks. Thus, our network benefits from the use of a bidirectional LSTM (bi-LSTM). The bi-LSTM consists of two separate LSTMs in which one is focused on the forward direction of the sequence (i.e. $1\dots T$) and the other models the temporal information in the reverse direction (i.e. $T\dots 1$). Finally, the outputs of the both LSTMs are concatenated to represent the temporal information in both directions. In this way, a bi-LSTM captures the long range temporal dependencies in forward-backward fashion. Sharma et al. \cite{sharma2015action} observed weighting the 3D CNN outputs through attention mechanism provides higher recognition accuracy. Inspired by this approach, we adapt Multi-Head attention \cite{vaswani2017attention} from machine translation problem to map the output of the Inception-Resnet-V2 CNN model to a weighted version of itself. Instead of using 3D CNN outputs as in \cite{sharma2015action}, the proposed model uses average pooled 2D feature maps from the Inception-Resnet-V2 model. As a result, it reduces the network size and parameters. We also experimentally found that our model benefits from Multi-Head attention-based temporal processing. In the next section, we provide the details of the adapted Multi-Head attention mechanism.            
\subsection{Attention Mechanism}
In general, all attention mechanisms maps input values $V$ to weighted representations using keys $K$ queries $Q$. As a result, values $V$ focus on more discriminatory features. In applications like machine translation where encoder-decoder style architectures are normally used, $K$ and $V$ are obtained from decoder while $Q$ is from the encoder. For, self-attention mechanisms which essentially calculates weighted representations of itself, $K$, $Q$ and $V$ is the same vector. In case of Multi-Head Attention \cite{vaswani2017attention}, the input vector is divided to a number of parts called heads. Attention mapping is carried out for each head separately and the heads are linearly concatenated in a weighted manner to keep the input dimension same as the output. This results in the output maps focusing on different sub-spaces of the input vector. Formally, Multi-Head attention \cite{vaswani2017attention} can be represented as:
\begin{equation}
    \text{Attention}(Q, K, V)  = \text{softmax}(QK^{T_r}/\sqrt{d_k})V
    \label{eq:eq7}
\end{equation}
\begin{equation}
    \text{MultiHead}(Q, K, V) = \text{Concat}(head_{1},\dots ,head_{h})W^o
    \label{eq:eq8}
\end{equation}
\begin{equation}
    \text{where }head_{i} = \text{Attention}(QW_{i}^Q,KW_{i}^K,VW_{i}^V)
    \label{eq:eq9}
\end{equation}
where $T_r$ represents the transpose of a given vector/matrix. In our case, the mechanism is applied in a self-attention manner. This implies $K=Q=V$. As a result, $W_{i}^Q = W_{i}^K = W_{i}^V \mathbf{\in }\mathbf{R}^{{D}\times {D}}$ and $W^o \in \mathbf{R}^{{hD}\times {D}}$. When this is applied to the RGB stream, the final dimension is $D=1536$, which is same as the output of the Inception-Resnet-v2 network. For pose network, $D=120$ and is the enriched feature length obtained from the concatenation of SEU and TEU. The number of attention heads is experimentally found to be 4 i.e. $h=4$. We are also able to show that the adapted Multi-Head attention mechanism increases the recognition accuracy for both the RGB and the fused posed streams. Here, $d_{k}=D/h$ is a scaling factor.
\subsection{Combined Model: Fusion of three streams}
Multiple streams in a given model are usually combined using either early fusion or late fusion. In case of the proposed model, we use a hybrid approach in which the early fusion is focused on the fusion of the SEU and the TEU (see Fig. \ref{fig:model}, and late fusion combines the body pose and the RGB stream. Moreover, the early fusion considers features, which are extracted from the same feature space (e.g. body pose) whereas in late fusion, the features are combined from separate feature space (Pose and RGB stream). Before the late fusion, features in both the RGB and pose stream are processed through the stream-specific self Multi-Head attention followed by a bi-LSTM. After the late fusion, a GAP (Global Average Pooling) and a Fully Connected (FC) layer is used for the activity classification, as shown in Fig. \ref{fig:model}. 
\section{Experimental Evaluations}
In order to evaluate the performance of the proposed network, we use three widely used datasets: 1) MSR daily Activity \cite{wang2012mining}, 2) NTU-RGBD \cite{shahroudy2016ntu} and 3) SBU Kinect interaction \cite{yun2012two}. All datasets also provide 3D pose sequences along with the RGB videos for each action. We use the standard accuracy metric in percentage in all of our evaluations. The MSR daily activity and NTU-RGBD datasets contain mostly single person activity and human-object interaction videos. NTU-RGBD dataset is one of the largest and most challenging datasets containing ~56K videos and 60 different activity classes. For NTU-RGBD dataset, we follow the standard evaluation protocol in \cite{shahroudy2016ntu}. The authors suggest two modes of validation: cross-subject (CS) and cross-view (CV). We validate our results in the CS mode, which is more challenging. The results in Table \ref{table:ntu_compare} indicate that the proposed model comfortably outperforms other existing models. Our model using RGB+Pose outperforms the best performed state-of-the-art model (RGB only in \cite{Baradel_2018_CVPR}) by 1\%. The difference in performance is due to the additional pose encoding using our SEU and TEU sub-module. Moreover, our approach (87.7\%) is significantly better than the PDA \cite{baradel2018human} (84.8\%) and DSSCA-SSLM \cite{shahroudy2017deep} (74.9\%) approaches that use both pose and RGB information. Moreover, using RGB only our approach (85.3\%) is significantly better than the C3D \cite{tran2015learning} (63.5\%) but inferior to the Glimpse Clouds \cite{Baradel_2018_CVPR} (86.6\%).

\begin{table}[!t]
\caption{Performance of our model and comparison to other state-of-the-art approaches on the NTU RGB+D dataset \cite{shahroudy2016ntu}. All the results are in cross subject settings which is more challenging than the cross view settings }
\label{table:ntu_compare}
\centering
\begin{tabular}[c]{ l c c c }
        	\hline
            \textbf{Methods} & \textbf{Pose} & \textbf{RGB} & \textbf{Acc} (\%)\\
            \hline
            Part-aware LSTM \cite{shahroudy2016ntu} & X & - & 62.9\\
            C3D \cite{tran2015learning} & - & X & 63.5\\
            DSSCA-SSLM \cite{shahroudy2017deep} & X & X & 74.9\\
            Synthesized CNN  \cite{liu2017enhanced} & X & - & 80.0\\
            ST-GCN \cite{yan2018spatial} & X & - & 81.5\\
            DPRL+GCNN \cite{tang2018deep} &  X & - & 83.5\\
            PDA \cite{baradel2018human} & X &  X & 84.8\\
            3Scale ResNet152 \cite{li2017skeleton} & X & - & 85.5\\
            Glimpse Clouds \cite{Baradel_2018_CVPR} & - & X & 86.6\\
            \hline
            Ours (Pose)  & X & - & 77.3\\
            Ours (RGB)   & - & X & 85.3\\
            Ours (Pose+RGB) & X & X & 87.7\\            
            \hline
		\end{tabular}
\end{table}

 The MSR Activity dataset \cite{wang2012mining} contains 320 videos with 16 different activity classes from 10 subjects. We follow the standard evaluation protocol \cite{wang2012mining} where the first 5 subjects are used for training and the remaining 5 are for validation. The performance using this dataset is presented in Table \ref{table:MSR_comparison}. The proposed model using  outperforms (92.5\%) the PDA approach \cite{Baradel_2018_CVPR} (90.0\%) using RGB+pose data. Using RGB only, our approach (90.6\%) is significantly better than all the approaches use uni-modal information. The best performing models use a combination of raw depth and pose data which is very memory intensive. Each MSR depth action consumes $45$ MB data while a RGB video requires only around $5$ MB. It is not feasible to scale such models to larger dataset. This indicates why many authors have ignored the raw depth based models in larger datasets like the NTU-RGBD.      
 \begin{table}[!t]
\caption{Comparison of proposed model with the state-of-the-art approaches on MSR dataset \cite{wang2012mining}}
\label{table:MSR_comparison}
\centering
\begin{tabular}[c]{ l c c c c}
        	\hline
            \textbf{Methods} & \textbf{Pose} & \textbf{RGB} & \textbf{Depth} & \textbf{Acc} (\%)\\
            \hline
            Ensemble \cite{wang2012mining} & X & - & - & 68.0 \\
            Efficient Pose \cite{eweiwi2014efficient} & X & - & - & 73.1 \\
            Moving Pose \cite{zanfir2013moving} &  X & - & - & 73.8 \\
            Poselets \cite{tao2015moving} & X & - & - & 74.5 \\
            MP \cite{shahroudy2017deep} & X & - & - & 79.4 \\
            Actionlet \cite{wang2013learning} & X &  - & - & 85.8\\
            PDA \cite{Baradel_2018_CVPR} & X &  X & - & 90.0\\
            \hline
            Depth Fusion \cite{zhu2015fusing}  & - & - & X & 88.8\\
            MMMP \cite{shahroudy2015multimodal} & X & - & X & 91.3\\
            DL-GSGC \cite{luo2013group}& X & - & X & 95.0\\
            DSSCA-SSLM \cite{shahroudy2017deep} & - & X & X & 97.5\\
            \hline
            Ours (Pose)  & X & - & - & 76.3\\
            Ours (RGB)   & - & X & - & 90.6\\
            Ours (Pose+RGB) & X & X & - & 92.5\\ 
            \hline
		\end{tabular}
		\vspace{-2mm}
		\label{table:MSR_compare}
\end{table}

The third dataset is SBU kinect \cite{yun2012two} interaction dataset, which is a human-human interaction datasets consisting of 282 videos with 8 different activities classes. For evaluation, we follow the authors' protocol of 5-fold cross-validation. The results in Table \ref{table:SBU_comparison} indicate that our model (96.5\%) significantly outperforms the state-of-the-art approaches that use either RGB or pose or their combination. Moreover, our model using pose only (96.2\%) is 2.9\% better than the best approach that uses ST-LSTM + Trust Gate \cite{liu2016spatio}.   

 \begin{table}[!t]
\label{table:SBU_comparison}
\centering
\caption{Results on the SBU Kinect dataset \cite{yun2012two}. The results shown are the average of 5 fold cross-validation}
        \begin{tabular}[c]{ l l l l l }
        	\hline
            \textbf{Methods} & \textbf{Pose} & \textbf{RGB} & \textbf{Depth} & \textbf{Acc} (\%)\\
            \hline
            Joint feature \cite{yun2012two} & X &  - & - & 80.3\\
            Joint feature \cite{ji2014interactive} & X &  - & - & 86.9\\
            Co-occurence RNN \cite{zhu2016co} & X &  - & - & 90.4\\
            STA-LSTM \cite{song2017end} & X &  - & - & 91.5\\
            ST-LSTM + Trust Gate \cite{liu2016spatio} & X &  - & - & 93.3\\
            DSPM \cite{lin2016deep} & - & X & X & 93.4\\
            PDA \cite{baradel2018human} & X &  X & - & 94.1\\
            VA-LSTM \cite{zhang2017view} & - & X & X & 97.5\\
            \hline
            Ours (Pose)  & X & - & - & 96.2\\
            Ours (RGB)   & - & X & - & 95.5\\
            Ours (Pose+RGB) & X & X & - & 96.5\\ 
            \hline
		\end{tabular}
		\vspace{-2mm}
		\label{table:SBU_Kinect}
\end{table}
\section{Implementation}
For all datasets, we sample a sub-sequence of 20 equally spaced frames. For all the pose data, a normalization step is applied where the data is transformed to body centered coordinates. This is done by subtracting the ``middle of spine'' joint from each joint and then normalizing with respect to the ``middle of spine'' joint. In case of multiple subjects, normalization is carried out on each subjects separately. The video sequences are cropped to a size of 224x224 and the pose sequences are translated accordingly. 
The model is trained using the Adam optimizer with a fixed initial learning rate of 1e-3 and a decay rate of 1e-6. However, while experimenting with the skeleton model, Stochastic Gradient Descent (SGD) optimizer has been used with learning rate of 0.1. The regularization factor is set at 1e-5 with $L_2$ regularization. The network has been trained on an Ubuntu PC fitted with an Nvidia Quadro P6000 (24 GB). Mini-batch sizes of 4 were used for 200 epochs to train the model and the categorical cross entropy is used as a loss function. The proposed model is implemented on Tensorflow with Keras wrapper.
%
%
In this section, we study the 
impact of the performance of individual components such as SEU, TEU and Multi-Head attention-mechanism in both RGB and Pose stream. We also analyze their positive impact on the network performance. For RGB stream, we use a pre-trained Incpetion-ResNet-V2 \cite{szegedy2017inception} followed by a bi-LSTM module as our base network. This is followed by a GAP and FC layer for training and evaluation. Later the Multi-Head attention mechanism is included to evaluate its effectiveness. As shown in table \ref{table:rgb_attention}, the attention mechanism significantly enhances the performance of the RGB stream as compared to the base network. This justifies the significance of the attention module in our network.
\begin{table}[ht]
    \centering
	\caption{Experiments show that application of Self Multi-Head attention mechanism to the RGB network improves the performance significantly. $+$ signifies the addition of that sub-module}
        \begin{tabular}[c]{ l l l l }
        	\hline
            \textbf{Method} & \textbf{NTU} & \textbf{MSR} & \textbf{SBU}\\
            \hline
            Baseline & 82.18 &  86.9 & 91.7\\
            + Multi-Head Self Attention & 86.6 & 90.6 & 95.5\\
            \hline
		\end{tabular}
		\vspace{-4mm}
		\label{table:rgb_attention}
\end{table}

In the base network, instead of the SEU and TEU sub-modules, we use three 1D convolution layers. It also does not include the Multi-Head attention mechanism. For evaluating the pose model, a Fully Connected (FC) layer is applied on top of the final output of the pose network. Afterwards, we have experimented by introducing the SEU in first three 1D convolutional layers of one of the stream. Then, in the second stream, the TEU is introduced in the first three 1D convolution layers. The Multi-Head self attention mechanism is applied to the pose network after the fusion of two streams. Experiments showed that keeping the number of heads at 4 is optimum for both RGB and pose networks. The table \ref{table:pose_network} shows considerable improvement as a result of SEU, TEU and the Multi-Head Attention mechanism. 
\begin{table}[ht]
    \caption{The performance of each network element. $+$ signifies the addition of that sub-module}
    \centering
        \begin{tabular}[c]{ l l l  }
        	\hline
            \textbf{Method} & \textbf{NTU} & \textbf{MSR} \\
            \hline
            Baseline & 73.3 &  72.5\\
            + SEU & 75.4 & 74.3\\
            + TEU  & 75.9 & 75.0\\
            + Multi-Head Self Attention & 77.3 & 76.3\\
            \hline
		\end{tabular}
		\vspace{-2mm}
		\label{table:pose_network}
\end{table}
\section{Conclusion}
In this article, we have proposed a novel method for learning enriched feature representation from 3D skeleton sequences instead of handcrafting such features. This representation is applied to our new multi-stream network that consists of two pose and one video streams. Out of the two pose streams, the first stream presents spatially enriched 3D pose data that captures the structural relationships between the various body joints and learns spatially enhanced representation. The second stream, learns the temporal relationship between various time points for each joint individually and presents a temporally enhanced representation. The proposed network uses a hybrid fusion approach in which two pose streams are combined using early fusion, and then the combined pose and RGB streams are fused using the late fusion. During fusion, we have also used Multi-Head self-attention mechanism that gives us state-of-the art results in three datasets.
%


\bibliographystyle{IEEEtran}
\bibliography{egbib.bib}
%

\end{document}